\documentclass[10pt]{article}
\usepackage[ansinew]{inputenc}
\usepackage{fancyhdr}

\usepackage{amssymb}
\usepackage{tipa}
\usepackage[ansinew]{inputenc}
\usepackage{fancyhdr}
\usepackage[all]{xy}
\usepackage{sectsty} 
\usepackage[normalem]{ulem}
\allsectionsfont{\large}
\subsectionfont{\normalsize}
\usepackage{amsmath}
\usepackage{mathenv}
\usepackage{dsfont}

\usepackage{graphicx}
\usepackage{color}
\usepackage[colorlinks=true, citecolor=blue, linkcolor=blue, urlcolor=blue]{hyperref}
\usepackage[active,new,noold,marker]{xrcs}
\usepackage[active]{srcltx} 

\title{\begin{flushright} \vspace{-70pt}\textit{\footnotesize Computer Science, Engineering \& \\ \vspace{-10pt}  General Sciences, Prop. Addendum} \\
\textit{}\\\vspace{-20pt}
\textit{\footnotesize  Comp. \& Phys.}$\,$\footnotesize \textbf{\textit{Vol.}}\textit{ }\textbf{1}, \emph{Ver.} 1, 1--10\\ \vspace{2pt} \footnotesize Postpub. Addendum, on 06 Nov 2007
\\ --------------------------------------------------- \end{flushright} \vspace{12pt} \Large  \textbf{Addendum to Research MMMCV; \\ A Man/Microbio/Megabio/Computer Vision}\\
}

\begin{document}


\definecolor{MyPurple}{rgb}{0.4,0.08,0.45}

\definecolor{MyBrown}{rgb}{0.6,0.4,0}
\definecolor{MyTeal}{rgb}{0.1,0.4,0.6}
\date{}
\maketitle
\begin{center}
\vspace{-50pt}
\long\def\symbolfootnote[#1]#2{\begingroup%
\def\thefootnote{\fnsymbol{footnote}}\footnote[#1]{#2}\endgroup}
\smallskip
By Philip Baback Alipour \noindent\symbolfootnote[1]{Author for correspondence (\htmladdnormallink {\textcolor{blue}{philipbaback\_orbsix@msn.com}}{mailto:philipbaback_orbsix@msn.com}).} $^{,}$ \noindent\symbolfootnote[2]{This paper represents the research proposal on MMMCV and biovielectroluminescence, Article-id:\,\htmladdnormallink{0710.0410}{http://arxiv.org/abs/0710.0410v1}\,[cs.CE], where the practical phase demonstrating the phenomenon in form of a research report, shall be published in arXiv.org and scientific journals.}
\end{center}

\begin{center}
\textit{Computer Science, Physics and Biology, Personalized Research Project,}

\textit{Elm Tree Farm, Wallingfen Lane, Newport, Brough, HU15 1RF, UK}



\end{center}

\noindent \textbf{Abstract ----- }\small {In October 2007, a Research Proposal for the University of Sydney, Australia, the author suggested that biovie-physical phenomenon as `\emph{electrodynamic dependant biological vision}', is governed by relativistic quantum laws and biovision. The phenomenon on the basis of `\emph{biovielectroluminescence}', satisfies man/microbio/ megabio/computer vision (MMMCV), as a robust candidate for physical and visual sciences. The general aim of this addendum is to present a refined text of Sections 1-3 of that proposal and highlighting the contents of its Appendix in form of a `Mechanisms' Section. We then briefly remind in an article aimed for December 2007, by appending two more equations into Section 3, a theoretical \textbf{II}-time scenario as a time model well-proposed for the phenomenon. The time model within the core of the proposal, plays a significant role in emphasizing the principle points on Objectives no.\,1-8, Subhypothesis\, 3.1.2, mentioned in Article [\htmladdnormallink{arXiv:0710.0410}{http://aps.arxiv.org/abs/0710.0410}]. It also expresses the time concept in terms of causing quantized time-energy function $f(|E_t|)$, emit in regard to shortening the probability of particle loci as predictable patterns of particle's \emph{un-occurred motion}, a solution to Heisenberg's uncertainty principle (HUP) into a simplistic manner. We conclude that, practical frames via a time algorithm to this model, fixates such predictable patterns of motion of scenery bodies onto recordable observation points of a MMMCV system. It even suppresses or predicts superposition phenomena coming from a human subject and/or other bio-subjects for any decision making event, e.g., brainwave quantum patterns based on vision. Maintaining the existential probability of Riemann surfaces of \textbf{II}-time scenarios in the context of biovielectroluminescence, makes motion-prediction a possibility.

\long\def\symbolfootnote[#1]#2{\begingroup%
\def\thefootnote{\fnsymbol{footnote}}\footnote[#1]{#2}\endgroup}
\begin{center} \small
\textbf {\footnotesize{Keywords: biovielectroluminescence;\noindent\symbolfootnote[3]{Pronounced as \textcolor{blue}{ \textipa{b\=i$'$\=o-v\=i-$\breve{\mathrm{i}}$-l$\breve{\mathrm{e}}$k"tr\=o-l\=u"m@-n$\breve{\mathrm{e}}$s$'$@ns}}, or see publication, \S1, Ref.~\cite{02-Alipour}.} special, general and unified relativity; Riemann surface; pixel velocity, uncured un-occurred motion }}
\vspace{0pt}
\smallskip
\end{center}
\section{Introduction}
\label{section1}
\vspace{1pt}
\markboth{}{Addendum to Research MMMCV}

\long\def\symbolfootnote[#1]#2{\begingroup%
\def\thefootnote{\fnsymbol{footnote}}\footnote[#1]{#2}\endgroup}

\normalsize I propose to design, construct and conduct the experiment by achieving the following goals in the project's course of implementation: \noindent\symbolfootnote[4]{The reader is reminded that such proposals are written for a very broad reader-ship, as the evaluators come not only from physics, but fields as far as biology, and computer sciences. The proposal was submitted by the current author in early November 2007 in his `Computer Science PhD Admission Application, University of Sydney'. The addendum's first three sections in content, coincide with the Sections 1-3 of the proposal.}

\begin{enumerate}
\item [{I.}] Present and explain, \textbf{i}- the theoretical presence of biovielectroluminescence via fly's vision, \textbf{ii}- the biovielectroluminescence phenomenon under laboratorial conditions via at least one prototype relative to a fly and its associated engineering modules, \textbf{iii}- pre/post-motion frame expectations on patterns of motion via biovielectroluminescence technology, e.g., a mountable visual + imaging unit on a man's head.

\item [{II.}] \textbf{iv}- Explain and prove how this new phenomenon unifies Einstein's special and general theories of relativity as claimed in terms of \textit{unified relativity} (UR)~\cite{08-Einstein et al.}. (See \S\ref{section2} for more details.)

\item [{III.}] \textbf{v}- Submit unequivocal evidence of image-data source and sink, giving compatible/convertible vision of a fly's vision into a human visual system.

\item [{IV.}] \textbf{vi}- Link all motion detection in climbing versus falling rates compared to a fly, prior to `mobile/stationary scenery objects' for human vision using this technology in practice.
\end{enumerate}

The above-mentioned objectives must be dealt with chronologically (I-IV). The detailed version of Objs.\,i-vi are Objs.\,1-8, pp.\,2-3, \S1, Ref.~\cite{02-Alipour}. The base plan, processes, anatomic features, components and materials, are illustrated in Figs.\,1 and 2. These illustrations are the exact duplicates of those figures on pp.\,9 and 25, Ref.~\cite{02-Alipour}. More detailed versions are not given in this proposal. (Refer to Pub.~\cite{02-Alipour}, and those relevant explanations provided on its figures.)

\section{Background}
\label{section2}
The core experimental investigation upon the project is interrelated and intermingled with the following publication:
\begin{center}
\textbf{The Theory }\textit{of}\textbf{ Unified Relativity }\textit{for a}\textbf{ Biovielectroluminescence Phenomenon }\textit{via}\textbf{ Fly's Visual and Imaging System} \\
\textit{}\\
By Philip Baback Alipour\\
\textit{}\\
                        Article-id: \htmladdnormallink{arXiv:0710.0410}{http://aps.arxiv.org/abs/0710.0410} [\textbf{cs.CE}] and [\textbf{cs.CV}], 2007\\
\end{center}

\noindent which strictly relates to subjects of \textit{Computational Engineering, Finance and Science}, including, \textit{Computer Vision and Pattern Recognition}. The publication is also an extension to a scientific book written on \textit{Advanced Physics} and \textit{Computing}, embodying my discoveries, new scientific projects and ideas into its structure~\cite[i-vii]{01-Alipour}.

\vspace{62.5mm}
\begin{flushleft}
\includegraphics[width=24mm, viewport= 0 0 10 50]{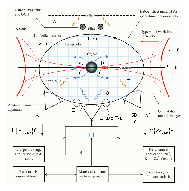}\\
\end{flushleft}
\vspace{-2pt}

\noindent \footnotesize{ \textbf{Figure 1. }The plan, the hardware/software engineering-involved processes and algorithm. An exact duplicate of Fig.\,3.1,~\cite{02-Alipour}.} Entire diagram by the current author. \\

\normalsize
The ending product of the research, based on the theoretical analysis and other aspects of the project, will be the following equations:
     \[ \ \ \ \ \ \ \ \ \ \ \ \ \ \ \ \ \ \ \ \ \ \ \ \ \ \ \mathbf{E}_{{\rm biovi}}^{L^{_{*} } } =\frac{VI}{A_{{\rm fly}\left(\sphericalangle \right)} } {\rm \; in\; Wm}^{-2} {\rm or,\; kg\:s}^{-3} , \ \ \ \ \ \ \ \ \ \ \ \ \ \ \ \ \ \ \ \ \ \ \ \ \ (1)\]
\noindent deriving
    \[ \ \ \ \ \ \ \ \ \ \ \ \ \ \ \ \  \ \ \ \ \ \ \ \ \ \ \ \ \  \beta _{\nu } =Ev_{r,g,b} {\rm \; in\; \; kg\:m}^{3} {\rm s}^{-3} {\rm \; or,\; \; J\:ms}^{-1} \ , \ \ \ \ \ \ \ \ \ \ \ \ \ \ \ \ \ \ \ \ \ \ \ \ (2) \]

\noindent which gives out a robust perspective on the physics and imagery dynamics of the project in action, whereby, I concluded in these equations, hypothetical formulations justifiably in the published article~\cite{02-Alipour}. Notation $\beta _{\nu } $ in Eq.\:(2), represents the phenomenon's direct expanded function, and $\textbf{E}_{{\rm biovi}}^{L^{_{*} } } $ in Eq.\:(3) the phenomenon's multifunction in \textbf{E}-field equations, both define biovielectroluminescence in physical measurements (pp.\,35-36, Ref.~\cite{02-Alipour}).
\vspace{62.5mm}
\begin{flushleft}
\includegraphics[width=27mm, viewport= 0 0 10 43]{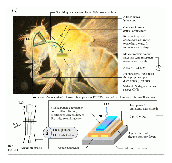}\\
\includegraphics[width=27mm, viewport= 0 0 10 21]{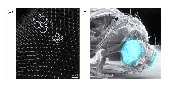}
\end{flushleft}
\vspace{-2pt}

\footnotesize \noindent \textbf{Figure 2.}  Anatomic features, hardware engineering components with their effects and etc. An exact duplicate of Fig.\,2.2,~\cite{02-Alipour}. Diagrams and pictorial modifications by the current author. \\

\normalsize
Eq.\,(2) components $E$, $v_{r,g,b} $, $V$, $I$ and $A_{{\rm fly}\left(\sphericalangle \right)} $, represent the phenomenon's energy, pixel velocity, voltage, electric current, fly's eye surface area,\:respectively. For the \textbf{E}-field concept, one must be familiar with Maxwell's equations~\cite[ii]{10-Wiki}.

\section{Background extension on MMMCV main formulae}
\label{section3}
\subsection {The II-time scenario and uncertainty-certainty relation}

For the time problem, it is contemplated for the energy part of $\beta _{\nu } $, $E$, a quantum relationship between $\beta _{\nu } $'s components in form of II-time scenarios representing quantum behaviour of Eqs.\,(1) and (2). For instance, Eq.\,(2) in II-time scenario algorithms, promotes into
    \[ \ \ \ \ \ \ \ \ \ \ \ \ \ \ \ \ \ \ \ \ \ \ \ \ \ \ \ \ \ \ \ \ \ \ \ \ \  \mathcal{E}_\emph{\textbf{t}}^2\equiv f(|E_t|) =\frac {[L,L']}{[t,t']} = \frac{|n m_j\hbar h| }{ S_t } = \frac{|\beta _{\nu }|^2}{|v_qv_{r,g,b}|}  \  \ \ \ \ \ \ \ \ \ \ \ \ \ \ \ \ \ \ \ \   (3) \]

\noindent where $\mathcal{E}_\emph{\textbf{t}}^2$ is the quantized energy of time through its displacing contour surface $S_t$ in a Riemann surface stack. Relativistically, the quantization is defined by absolute function $f(|E_t|)$ in expansion to its quantized vectors in space-time UR relationships. Notation $L$ and its symmetry $L'$, represent angular momenta of particle of charge $q=-e$, where $-e$ is the electron relative to angular momentum surfaces, possessing time $t$ against $t'$ of some scenery body's clock (e.g., a human subject's clock), respectively. Both momenta are intermingled to quantized angular momentum whilst performing orbital angular momentum for the particles across the stack. The quantized relation between space-time events are in form of $\sqrt{z}$ product, $\sqrt{z}\cdot\sqrt{z}$, relevant to the Riemann stack formation.

This stack formation, forms magnitudes of principle quantum number $n$ multiplied by magnetic quantum number $m_j$, Planck's constant $h$ and Dirac's constant $\hbar$, thereby divided by contour surface quantity $S_t$. This equivalently generates by definition, quantized $\beta _{\nu }$ divided by a magnitude of particle's velocity $v_q$ multiplied by pixel velocity $v_{r,g,b}$, in process to Eq.\,(1). The last equations succeed to the subsequent \emph{uncertainty-certainty relation} signifying the importance of wave propagation and its distribution factor in Fig.\,2.3 of Ref.~\cite{02-Alipour} with relevance to Eqs.\,(2.4) to (2.5), now in form of
\[ \ \ \ \ \ \ \ \ \ \ \ \ \ \ \ \ \ \ \  \ \ \ \ \ \ \ \ \ \ \ \Delta x_k \Delta p_q \{<,=\}\,\left|\, \{=,>\} \frac {\hbar}{2}\right. \stackrel {\frak A}{\longrightarrow} \Delta x \Delta p > \ngtr \frac{\hbar}{2} \ \ \ \ \  \ \ \ \ \ \ \ \ \ \ \ \ \ \ \ \    (4)\]
\long\def\symbolfootnote[#1]#2{\begingroup%
\def\thefootnote{\fnsymbol{footnote}}\footnote[#1]{#2}\endgroup}
\normalsize
\noindent where operation conditions' set $\{<,=\}$ against set $\{=,>\}$, satisfying `curvature $k$-possessed length change' $\Delta x_k$, and particle momentum $p_q$, are relativistically provable in shortening particle's loci in displacement, solely when, $\Delta x$ is described as a distance in consumption or fragments of $\Delta x$ in anticipation (see the following  illustration). This consuming type of length via some anticipation function $\mathfrak A$ over \emph{unoccured events},\symbolfootnote[1]{Alternatively, `\emph{un-oc-cured}' when events are not processed yet or \emph{uncured}, and remain hypothetical. See parag.\,1, \S\ref{section5}, and the same word sporadically used by the author in Pub.~\cite{02-Alipour}.} is now consumed distance $x_{\mathrm{cons}}$ on $\Delta x$ from relation (2.1b) Ref.~\cite{02-Alipour}, which thus predicts particle loci against regular expectation of HUP-preserving conditions of $\Delta x \Delta p\ge \hbar / 2$ on the righthand side of the given restriction (denoted by the vertical bar symbol,`$|$', between the sides). In contrast, the lefthand side preserves conditions of $\{<,=\}$ when $\Delta x - \Delta x_k$ approaches 0 (see the right side of Fig.\,3). The consumption also fulfills the notion of `\emph{transformation of wave’s unit circle paradigm}' propounded on pp.\,13-17,~\cite{02-Alipour}. Operators $>$ and $\ngtr$ represent the above-noted operators in sets, in favour of \emph{uncertainty-certainty characteristics}, submitted in Fig.\,3.

\begin{flushleft}
\includegraphics[width=20.5mm, viewport= 0 0 10 37]{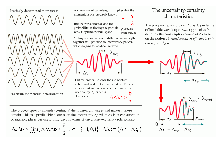}
\end{flushleft}
\vspace{-2pt}

\noindent \footnotesize \textbf{Figure 3.} An illustration on the characteristics of predictable position and momentum of a particle in scenery space in the context of biovielectroluminescence and MMMCV's bendable wavelength $\lambda_{\mathrm{bend}}$ inclusions. Localized wavelength distribution satisfies predictable chances on particle loci, approaching levels of \emph{certainty} in contradiction with HUP. Left side, courtesy of R. Nave, \emph{Dept. of Phys. \& Astro.}, Georgia State University; right side, by the author.

\subsection {Hypothetical outcomes}
\normalsize
Functional product Eq.\,(3) and \emph{uncertainty-certainty} Rel.\,(4), clarify the speciality of the project outcomes such as, postulates and hypotheses, and is subject to an article aimed for December 2007, on behalf of the progressing research investigating time problems to their quantum physical relationships in release.
\vspace{-6pt}
\section{Significance}
\label{section4}
\begin{enumerate}
\item [{1-}] This project detects \emph{un-occurred motion} of some moving body before its pattern being formed in action, framed into its kinematics via its \textit{bendable wavelength} \textit{function}, $\lambda _{\mathrm{bend}} $, pp.\,13-16, \S2, in recognition of its dynamics via function $\beta _{\nu } $ of Eq.\,(2) defined initially in \S5, Ref.~\cite{02-Alipour}.

\item [{2-}] The project performs a new \textit{bio-vi-physical} \textit{phenomenon} governed by quantum laws and biovision. I call the phenomenon on the basis of pp.\,2-6, biovielectroluminescence, which satisfies MMMC Vision, promising scientific awards; Candidacy for physical and visual sciences. (See also \S\ref{section5}.)

\item [{3-}] The phenomenon and the fulfilled Objs.\,\,i-vi on p.\,2, submit \textit{n}-dimensional vision and clocking system(s), detecting motion much faster than a human eye-`not benefiting' from biovielectroluminescence technology (the mountable device mentioned in Obj.\,iii). This is \underbar{very significant} for security systems and law enforcement that highly depend on image data.
\end{enumerate}

\noindent Other aspects of the project's usage are in e.g., military, medicine, holography, 3D scanning techniques, 360$^\circ$ vision and etc., also given in the Abstract and Conclusions Sections of my paper~\cite{02-Alipour}. Products developed from this project could be a robust substitution for, e.g., Fish Eye lens technology, due to being inexpensive in mass production for markets. (See \S\ref{section6}.) Researches that relate to this topic are done in a few occasions which solely cover \emph{real-time pattern recognition} on some body's motion. In contrast, this project gives an observer the potential grounds to detect \textit{un-occurred motion} of some body (\textit{motion pattern prediction}) in its forming path, which is on its own, \underbar{very significant}. To this account, based on the project's obvious grounds, the project will enhance and advance research topics e.g., \textit{Car Plate Recognition} and \textit{Human Motion Tracking} \textit{in Video}, as generalized on Prof. Hong Yan's website~\cite{06-Yan}.

\section{Operation Mechanism and Algorithm}
\label{section5}
This section gives a roughly accurate notion of algorithmic expectable outcomes in a hypothetical manner based on MMMCV (p.\,6, Significance\,\#2). The original report based on real figures and data representations are the aim of the actual project. The I/O algorithmic representation of predictable \emph{uncured motion} as a \textit{hypothetical exemplar} is shown in Fig.\,4. The notion of \emph{\textcolor{MyTeal}{uncured} \textcolor{MyBrown}{un}-\textcolor{MyPurple}{occurred} \textcolor{MyBrown}{motion}} (in short, \emph{\textcolor{MyBrown}{un}-\textcolor{MyPurple}{oc}-\textcolor{MyTeal}{cured} \textcolor{MyBrown}{motion}}) to the algorithm, could be interpreted in terms of \emph{a still unavailable motion on visual memory bank} or in brief, \emph{unprocessed and unrecorded motion pattern}. In this algorithm, in the multi-lens matrix frames, the superposition phenomenon coming from a human subject and/or other bio-subjects for any decision making event (brainwave quantum patterns based on vision), are suppressed or predicted due to Riemann surfaces,~\cite[v]{10-Wiki}, of II time scenarios in the context of biovielectroluminescence. (The theoretical report is subject of my next publication in December 2007.)

Notation $\mathbf{E}_{{\rm biovi}}^{L^{_{*} } } $ incorporated in Fig.\,4 algorithm, is of Eq.\,(1) on p.\,3; Symbols $t_{{\rm current}} $, $t_{{\rm future}} $, $t_{{\rm predicted}} $ denote current, future and predicted time events of mobile bodies, respectively. For the expression, $t_{cons} <t-\varepsilon $, see, Eq.\,(2.1a), and the explanations provided on p.\,6 of Ref.~\cite{02-Alipour}. In Fig.\,4, the \textit{prediction layers} are classified as negative, $-$ , indicating motion event's hidden layer(s) of prediction, which if predicted, is a \textit{successful prediction}; positive , $+$ , a \textit{failed prediction}; zero, 0, the start-to-closure of a new prediction cycle, otherwise, its fall. One could describe the latter as \textit{prediction loop decision point}, denoting outputs of HALT otherwise, START. Note that, an observation scope in the algorithm in terms of, `observation scope $\in \mathbf{E}_{{\rm biovi}}^{L^{_{*} } } $', as a multifunction agent dependent, $\mathbf{E}_{{\rm biovi}}^{L^{_{*} } } $, here, for a fly's vision relative to scenery bodies in motion, is of Eqs.\,(5.2), (5.3), of Ref.~\cite{02-Alipour}, thus defined through principles of biovielectroluminescence.

\vspace{62.5mm}
\begin{flushleft}
\includegraphics[width=15.4mm, viewport= 0 0 10 30]{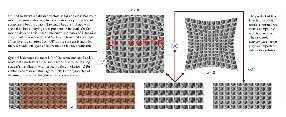}\\
\includegraphics[width=15.4mm, viewport= 0 0 10 49.5]{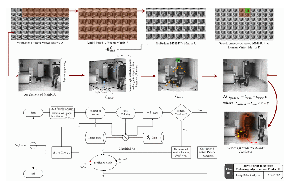}
\end{flushleft}
\vspace{-2pt}

\noindent \footnotesize \textbf{Figure 4.}  Expected human motion pattern process and analysis by an I/O pictorial and dataflow algorithm of Image Data based on MMMCV. In the final frame output (Element(s) of Matrix A-to-D, Predicted), the red motion path prediction is for the walking human and the yellow path, for the dog tagging along with the human subject. Other crucial elements such as `light' in form of energy \textit{h$\nu $}, or photon being projected onto the scenery subjects in, e.g., Matrix B, and reflection patterns denoting bendable wavelength $\lambda _{\mathrm{bend}} $, are subjects of Fig.\,1, p.\,3; and pp.\,13-16, \S2,~\cite{02-Alipour}. The two-frame photos, courtesy of Dr. Holden~\cite{11-Holden}; Image modifications, warping, the full algorithm and fractal pictures, by the current author. \\

\normalsize
Notation $\Delta t$ in one of the algorithm's process components, relates to a simplified `proper time' equation from Special Relativity, where $\Delta $ means `the change in' between two space-time events~\cite[iii and iv]{10-Wiki}. It is speculated that this change satisfies `II time scenarios' generating energy $E$, for a local time experienced on particle of charge $q=-e$'s clock in biovielectro-field of $\beta _{\nu } $ in Eq.\,2. (See, p.\,5, 1st parag.) In consequence, a unified relativistic space-time layering solution (Matrix C through D) is in the algorithm for a human vision product, outputting Matrix D.

\section{Conclusion and brief remarks}
\label{section6}
\vspace{4pt}

The proposed research culminates in a formal report once the funding support from the honorable academic body is in position, hence completing the research on its progressing grounds. Experimental analysis, spreadsheets and data reports would be based on Les Kirkup's experimental methods~\cite{07-Kirkup}. The main target of the basic experimental model is to exhibit 1- biovielectroluminescence, 2- pre-pattern recognition, and 3- motion analysis. The prototype would basically comprise of devices specifically designed to fulfill the requirements of Objs.\,1-8, Ref.~\cite{02-Alipour}, such that all probable constraints and data errors shall be provided in the report's supplements. These devices may incorporate Luke Lee's `Biologically Inspired Artificial Compound Eyes' depicted in Fig.\,2.1 of Ref.~\cite{02-Alipour}.

The general design also associates basic array of electro-optical units and detectors to conduct the three points made above. Once the actual experiment takes place, a database and software program module coded in VB or C++ to the computer system, thereby stores and analyzes scenery events as they occur on the observation scale and analysis stage. The software program is based on product frames of biovielectroluminescence, Sects. 4, 5,~\cite{02-Alipour}, and gives a general insight on industrial scale inexpensive mass production of MMMCV units in global markets. For satisfactory data results whilst considering cost reduction in the process, miniaturizing the complete model is deemed vital for this project.

In conclusion, this robust application recognizes systematic predications on microscopic against macroscopic patterns, scanning motion in any space-time situation for human vision synchronized to other visual systems, e.g., computer vision, insect vision in diversity.

\begin{flushleft}
\noindent \textbf{Statement of conclusion:} \end{flushleft}
\vspace{4pt}
\noindent --------- \emph{Assume in the laws of quantum physics, any assigned value as an input bring about the notion of `uncertainty'. Contradictorily, conjecturing pre-defined time and space in entanglement as `predetermined on anticipated space-time' for uncertain values in shortened intervals, is itself upon the characteristics of subatomic structures versus massive bodies to one's vision, a `certainty'} ---------

\begin{flushright}
The Author \end{flushright} \vspace{-12pt}

\section*{Acknowledgments}
\label{section7}
\vspace{4pt}
The author thanks G. E. Goodwin, External Examiner \emph{of} Leeds Metropolitan University, M. Dickinson, Former Senior Lecturer \emph{of} Mathematics, University \emph{of} Lincoln, for their written character reference support on behalf of the author's personalized scientific activities. It is highly appreciated for Dr. H. Alipour \emph{et al.}, on their moral support on the author's research-based endeavours.
\vspace{-2pt}
\textbf{  }


\begin{thebibliography}{99}
\markboth{}{References}
\vspace{6pt}
\footnotesize
\bibitem{01-Alipour} P. B. Alipour, \textit{The Compulsory Force on force Principia}, 370 pages, Copyrighted\copyright\,2007, license no:\:TXU001347562, including its demo software package (PhiBAl's Demo …) is a Copyrighted © and Registered ® property of the above-named author, years (1998-2007), License no:\,TXU001347596; PAU003133514, Library of Congress, Washington D.C., USA, \emph{it consists of III discoveries}; \emph{Its potential articles are}: \textbf{i-} The Principle of Parallel Relativity, \textbf{ii-} The Parallel Relativity's principal product: regular $\pi $-twist of multi-singularities' $\pi $-twist force system, \textbf{iii-} The III Scenario Matter and Parallel Relativity as; $\pi $-swirl twists relativity, \textbf{iv-} The 5th fundamental force, as a compulsive interacting-force of nature; the Compulsory Force on \textit{force}, \textbf{v-} The principle of time event tournament and synchronized reference frames, \textbf{vi-} The Planck's mutation of the Compulsory Force on \textit{force} system, \textbf{vii-} The Expansion of Einstein's Law, (1998-2007).

\bibitem{02-Alipour} P. B. Alipour, `The Theory of Unified Relativity for a Biovielectroluminescence Phenomenon via Fly's Visual and Imaging System', Article-id. arXiv:\,\htmladdnormallink{0710.0410}{http://arxiv.org/abs/0710.0410v1}, \textit{Comp. Sci., Comp. Eng. \& Vis.}, pp. 1-51, 2007.

\bibitem{03-Alipour} P. B. Alipour, `Theoretical Engineering and Satellite Comlink $\emph{of}$ a PTVD-SHAM System', arXiv:\,\htmladdnormallink{0710.0244}{http://arxiv.org/abs/0710.0244}, \textit{Comp. Sci., Comp. Eng. \& Ar. ArXiv.org.}, pp. 1-50, 2007.

\bibitem{04-Alipour} P. B. Alipour, `Logic, Design and Organization of Parallel Time Varying Data and Time Super-helical Memory As; PTVD-SHAM', Article-id. arXiv:\,\htmladdnormallink{0707.1151}{http://arxiv.org/abs/0707.1151}, \textit{Comp. Sci., Ar.,} \textit{ArXiv.org.}, pp. 1-33, 2007.

\bibitem{05-Alipour} P. B. Alipour, \textbf{Research Topic:} `The Biovielectroluminescence Phenomenon via Fly’s VIS for MMMCV; A Man/Microbio/Megabio/Computer Vision', \textit{Res. Prop.,} \textit{Comp. Sci., University of Sidney}, pp. 1-6, 14 Oct 2007.

\bibitem{06-Yan} Prof. H. Yan, Research Topics, \textbf{Research Areas:\textit{ }}Image processing, Pattern recognition, Bioinformatics \htmladdnormallink{http://www.ee.cityu.edu.hk/\~{}hpyan/pr\_im1.htm}{http://www.ee.cityu.edu.hk/\~hpyan/pr\_im1.htm}, Accessed on 06 Oct. 2007

\bibitem{07-Kirkup} L. Kirkup, \textit{Experimental Methods: an introduction to the analysis and presentation of data}, John Wiley \& Sons Press (1994) ISBN 0-471-33579-7

\bibitem{08-Einstein et al.} A. Einstein, H. A. Lorentz, H. Weyl and H. Minkowski, \textit{The principle of relativity}  pp. 88 and 90, 1952

\bibitem{09-Lee} L. P. Lee, J. Kim and K. H. Jeong, `Biologically Inspired Artificial Compound Eyes', \textit{J. Sci.}, {\textit{Vol.}} \textbf{312} (5773), pp. 557-561, 2006 (\htmladdnormallink {\textcolor{MyBrown}{doi:10.1126/science.1123053}}{http://dx.doi.org/10.1126/science.1123053})

\bibitem{10-Wiki} \textit{\htmladdnormallink{Online encyclopedia} {http://en.wikipedia.org/wiki/Main_Page}} (2007) \\ \textbf{i- }Electric Field on \htmladdnormallink{http://en.wikipedia.org/wiki/Electric\_field}{http://en.wikipedia.org/wiki/Electric_field} or, see, for instance \\ \textbf{ii-} Maxwell's equations on \htmladdnormallink{http://en.wikipedia.org/wiki/Maxwell\_equations}{http://en.wikipedia.org/wiki/Maxwell_equations} , \\ \textbf{iii-} Proper time on  \htmladdnormallink{http://en.wikipedia.org/wiki/Proper\_time}{http://en.wikipedia.org/wiki/Proper_time} , \\ \textbf{iv-} \textit{Basic concepts}, spacetime on   \htmladdnormallink{http://en.wikipedia.org/wiki/Spacetime}{http://en.wikipedia.org/wiki/Spacetime} , \\
    \textbf{v-} Riemann surface on \htmladdnormallink{http://en.wikipedia.org/wiki/Riemann\_surface}{http://en.wikipedia.org/wiki/Riemann_surface}.

\bibitem{11-Holden} E. Holden, `Detecting Humans in Video Footage using Multiple Classifiers' on \htmladdnormallink{http://www.csse.uwa.edu.au/\~{}eunjung/humandetect.html}
    {http://www.csse.uwa.edu.au/\~eunjung/humandetect.html}, Accessed on 07 Oct. 2007

\bibitem{12-AK} D. V. Ahluwalia-Khalilova, `Neutrino oscillations and supernovae', \textbf{a research proposal addendum example}, arXiv:astro-ph/0404055v1, 2 Apr 2004.

\bibitem{13-Zarate} J. P. McEvoy and O. Zarate,  \textit{Introducing Quantum}, `Heisenberg's gamma-ray microscope: de Broglie/Einstein relation \dots',  pp. 156, 158, 2003.

\bibitem{14-Dickinson} M. Dickinson and G. E. Goodwin, `Mathematics I \& II, \textit{Discrete Mathematics}', University of Lincoln, UK. (2002-2005).

\end{thebibliography}
\end{document}